\RequirePackage{fix-cm}
\documentclass[twocolumn]{svjour3}          % twocolumn
\smartqed  % flush right qed marks, e.g. at end of proof
\usepackage{graphicx}
\usepackage{subfigure}
\usepackage{rotating}
\usepackage{amsmath}
\usepackage{amsfonts}
\usepackage{url}
\usepackage{amssymb}
\usepackage{color}
\usepackage[round,authoryear]{natbib}
\usepackage{lipsum,afterpage}

\newcommand{\mytoppar}[1]{\vspace{-0mm}\paragraph{#1}}
\newcommand{\mypar}[1]{\vspace{-3mm}\paragraph{#1}}

\journalname{International Journal of Computer Vision Special Issue : BMVC 2018}
\begin{document}

\title{The Devil is in the Decoder: Classification, Regression and GANs}
%\title{Insert your title here%\thanks{Grants or other notes
%about the article that should go on the front page should be
%placed here. General acknowledgments should be placed at the end of the article.}
%}
%\subtitle{Do you have a subtitle?\\ If so, write it here}

%\titlerunning{Short form of title}        % if too long for running head

\author{Zbigniew Wojna \and 
Vittorio Ferrari \and 
Sergio Guadarrama \and \\
Nathan Silberman \and 
Liang-Chieh Chen \and 
Alireza Fathi \and  \\
Jasper Uijlings
}

\authorrunning{Wojna et al.} % if too long for running head

\institute{
Zbigniew Wojna \at University College London \\
\email{zbigniewwojna@gmail.com} \and
Vittorio Ferrari, Sergio Guadarrama, Nathan Silberman, \\
Liang-Chieh Chen, Alireza Fathi, Jasper Uijlings \at Google Inc. \\
\email{vittoferrari@google.com, sguada@google.com, nsilberman@google.com, lcchen@google.com, alirezafathi@google.com, jrru@google.com}
}

\date{Received: date / Accepted: date}
% The correct dates will be entered by the editor

\maketitle

\begin{abstract}
Many machine vision applications, such as semantic segmentation and depth prediction, require predictions for every pixel of the input image. Models for such problems usually consist of encoders
which decrease spatial resolution while learning a high-di\-men\-sional representation, followed by
decoders who recover the original input resolution and result in low-dimensional predictions. While
encoders have been studied rigorously, relatively few studies address the decoder side. 
This paper presents an extensive comparison of a variety of decoders for a variety of pixel-wise
tasks ranging from classification, regression to synthesis. Our contributions are: (1) Decoders
matter: we observe significant variance in results between different types of decoders on various
problems. (2) We introduce new residual-like connections for decoders. (3) We introduce a novel
decoder: bilinear additive upsampling. (4) We explore prediction artifacts.

\keywords{Machine Vision \and Computer Vision \and Neural Network Architectures \and Decoders \and 2D imagery \and per-pixel prediction \and semantic segmentation \and depth prediction \and GANs}
% \PACS{PACS code1 \and PACS code2 \and more}
% \subclass{MSC code1 \and MSC code2 \and more}

\end{abstract}

%\input{tempfile}

%-------------------------------------------------------------------------
\section{Introduction}\label{sec:intro}

Many important machine vision applications require predictions for every pixel of the input image.
Examples include but are not limited to: semantic segmentation (e.g.~\cite{long2015fully,tighe10eccv}), boundary
detection (e.g.~\cite{arbelaez11pami,uijlings2015situational}), human keypoints estimation (e.g.~\cite{DBLP:journals/corr/NewellYD16}), super-resolution
(e.g.~\cite{DBLP:journals/corr/LedigTHCATTWS16}), colorization (e.g. \cite{iizuka2016let}), depth estimation
(e.g.~\cite{Silberman:ECCV12}), normal surface estimation (e.g. \cite{DBLP:journals/corr/EigenF14}), saliency
prediction (e.g.~\cite{pan2016shallow}), image generation with Generative Adversarial Networks (GANs)
(e.g.~\cite{goodfellow14nips,nguyen2016plug}), and
optical flow (e.g.~\cite{ilg2016flownet}). Modern CNN-based models for such applications are usually
composed of a feature extractor that decreases spatial resolution while learning high-dimensional
representation and a decoder that recovers the original input resolution.

Feature extractors have been rigorously studied in the context of image classification, where
individual network improvements directly affect classification results. This makes it relatively
easy to understand their added value. Important improvements are convolutions \citep{lecun89nc},
Rectified Linear Units~\citep{nair10icml}, Local Response Normalization~\citep{krizhevsky12nips}, Batch
Normalization~\citep{ioffe15icml}, and the use of Skip Layers~\citep{bishop95,ripley96} for Inception
modules~\citep{DBLP:journals/corr/SzegedyVISW15}, ResNet~\citep{DBLP:journals/corr/HeZRS15}, and
DenseNet~\citep{huang17cvpr}.

In contrast, decoders have been studied in relatively few works. Furthermore, these works
focus on different problems and results are also influenced by choices and modifications of the
feature extractor. This makes it difficult to compare existing decoder types.
Our work, therefore, presents an extensive analysis of a variety of decoding methods on a broad range
of machine vision tasks. For each of these tasks, we fix the feature extractor which allows a direct
comparison of the decoders. In particular, we address seven machine vision tasks spanning a
classification, regression, and synthesis:
\begin{itemize}
\item Classification
  \begin{itemize}
  \item Semantic segmentation
  \item Instance edge detection
  \end{itemize}
\item Regression
  \begin{itemize}
  \item Human keypoints estimation
  \item Depth prediction
  \item Colorization
  \item Super-resolution
  \end{itemize}
\item Synthesis
  \begin{itemize}
  \item Image generation with generative adversarial networks
  \end{itemize}
\end{itemize}

We make the following contributions: (1) Decoders matter: we observe significant variance in results
between different types of decoders on various problems. (2) We
introduce residual-like connections for decoders which yield improvements across decoder types. (3) We propose a new bilinear additive
upsampling layer, which, unlike other bilinear upsampling variants, results in consistently good
performance across various problem types. (4) We investigate prediction artifacts.

\section{Decoder Architecture} \label{sec:method} 

\begin{figure}
\begin{center}
\includegraphics[width=0.8\hsize]{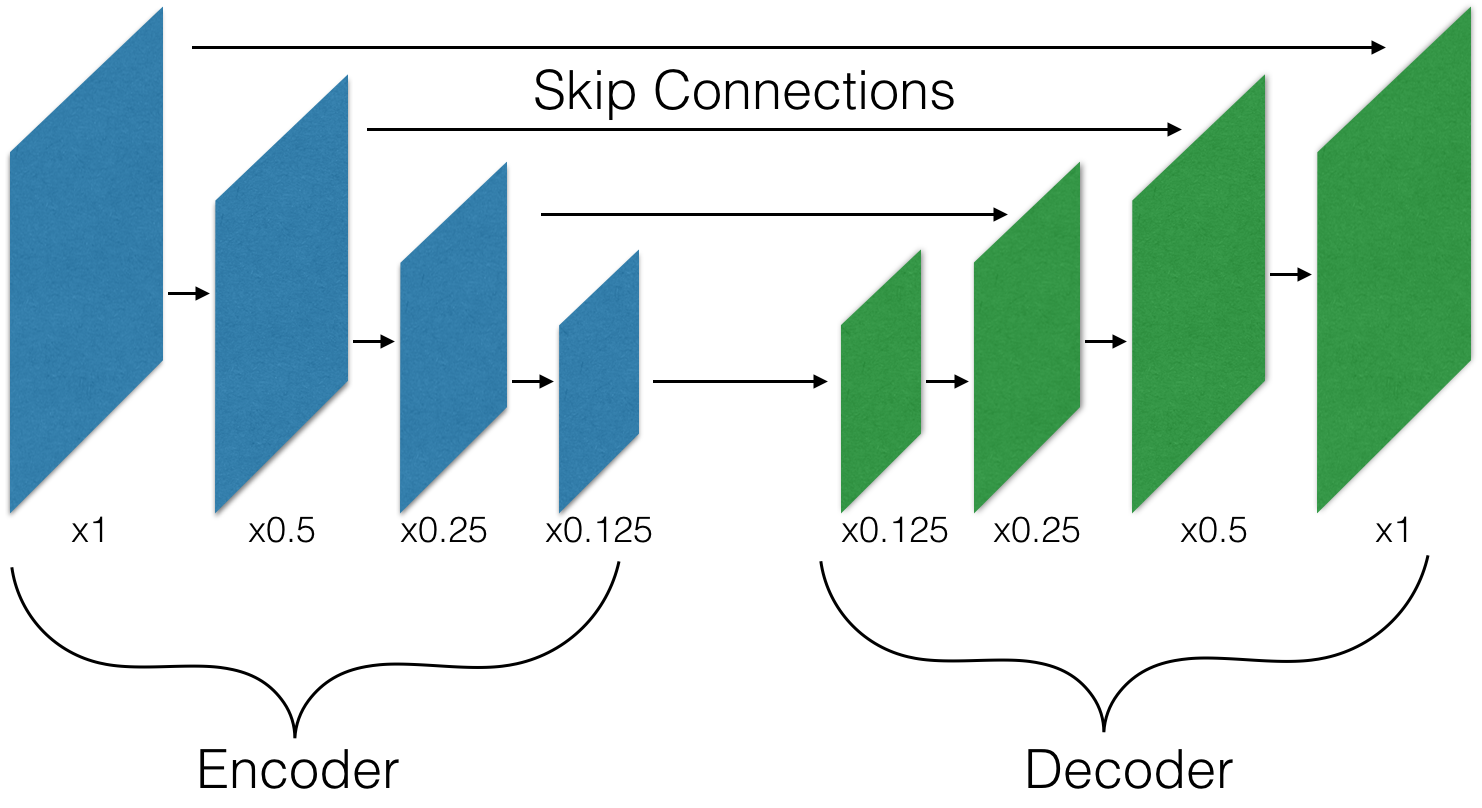}
\caption{General schematic architecture used for dense prediction problems. Typically, the resolution is reduced by a factor 2 in each of several encoder steps, after which they are upsampled by a factor 2. In this illustration, the lowest resolution is $0.125 \times$ the input size.}
\label{fig:encdec}
\end{center}
%\vspace{-.8cm}
\end{figure}

Dense problems which require per pixel predictions are typically addressed with an encoder-decoder
architecture (see Fig.~\ref{fig:encdec}). First, a feature extractor downsamples the spatial
resolution (usually by a factor 8-32) while increasing the number of channels. Afterwards, a
decoder upsamples the representation back to the original input size.  Conceptually, such decoder
can be seen as a reversed operation to what encoders do.  One decoder module consists of at
least one layer that increases spatial resolution, which is called an upsampling layer, and possibly
layers that preserve spatial resolution (e.g., standard convolution, a residual block, an inception
block).

The layers in the decoder which preserve spatial resolution are well studied in the literature in the
context of neural architectures for image classification (e.g., \cite{DBLP:journals/corr/SzegedyVISW15,
DBLP:journals/corr/Chollet16a, DBLP:journals/corr/AlvarezP16, DBLP:journals/corr/HeZRS15}). Therefore,
we only analyze those layers which increase spatial resolution: the upsampling layers.

\subsection{Related work}

Many decoder architectures were previously studied in the context of a single machine vision task.
The most common decoder is a transposed convolution, discussed in detail
in~\citep{dumoulin2016guide}, used for feature
learning~\citep{zeiler2010deconvolutional,zeiler11iccv} and pixel-wise prediction tasks such
as image segmentation~\citep{DBLP:journals/corr/RonnebergerFB15}, semantic
segmentation~\citep{long2015fully}, optical flow~\citep{fischer2015flownet}, depth
prediction~\citep{laina2016deeper}, image reconstruction from its feature
representation~\citep{dosovitskiy2016inverting}, and image synthesis~\citep{dosovitskiy15cvpr}.

Several convolution variants were proposed which trade model capacity for speed.  In
particular, in~\citep{DBLP:journals/corr/SzegedyVISW15} and \citep{romera18tits}
a 2D convolution is decomposed into two 1D convolutions in the context of image classification
and semantic segmentation.
In~\citep{DBLP:journals/corr/Chollet16a} separable convolutions are applied per channel to image
classification. We test both variants in the context of transposed convolutions.

Bilinear and/or bi-cubic upsampling were studied in the context of super-resolution by~\citep{dong16tpami}.  The depth-to-space upsampling method was proposed by ~\citep{shi2016real} to
improve the task of super-resolution. Our paper compares all these decoders, adds
one extra, and explores modifications, which are detailed in Sec.~\ref{sec:upsampling}.

A previous study by \citep{odena2016deconvolution} examined transposed convolution and bilinear
upsampling qualitatively in the context of GANs~\citep{goodfellow14nips}. We provide quantitative results comparing many more
decoders on seven different machine vision tasks.

Recently, stacked hourglass networks were proposed \citep{DBLP:journals/corr/NewellYD16}, which are
multiple encoder-decoder networks stacked in sequence. We include stacked hourglass networks in our analysis.

\subsection{Upsampling layers}\label{sec:upsampling}
Below we present and compare several ways of upsampling the spatial resolution in convolution neural networks, a crucial part of any decoder. We limit our study to upsampling the spatial resolution by a factor of two, which is the most common setup in the literature (see e.g.~\cite{DBLP:journals/corr/RonnebergerFB15} and \cite{yu2016ultra}). 

\subsubsection{Existing upsampling layers}
\begin{figure*}
  \begin{minipage}{0.5\textwidth}
  \centering
  \includegraphics[width=\hsize]{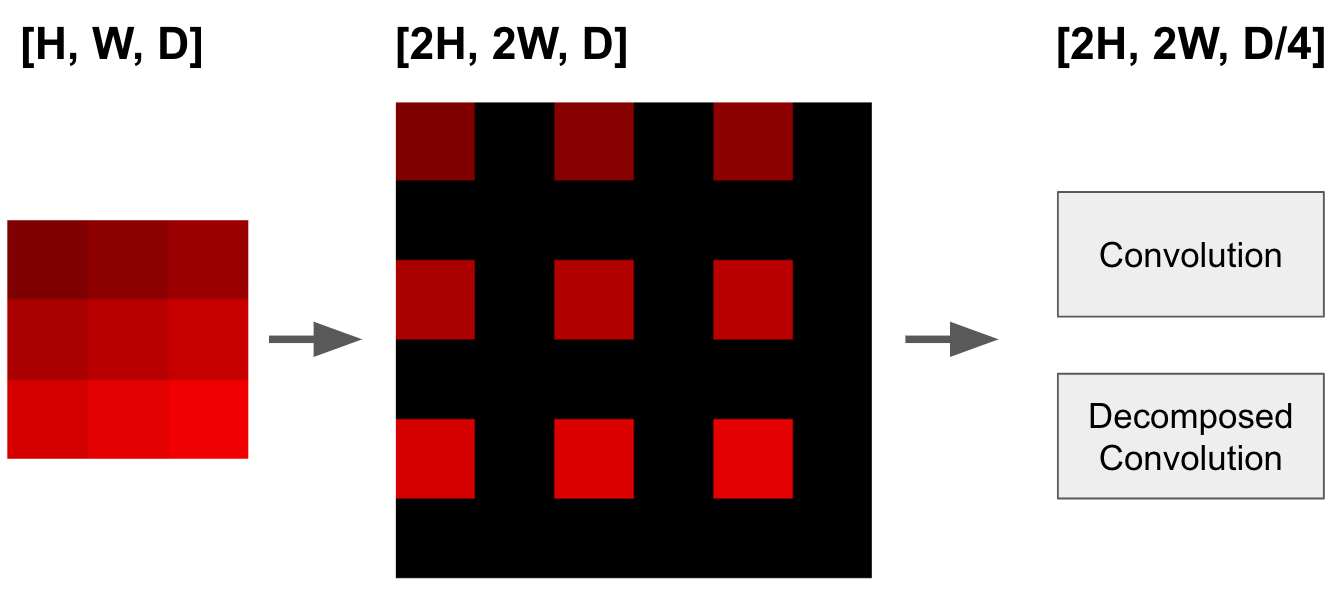}
  \caption{Visual illustration of (Decomposed) Transposed Convolution. The resolution is increased by interleaving the features with zeros. Afterwards either a normal or decomposed convolution is applied which reduces the number of channels by a factor 4.}
  \label{fig:transpose_visual}
  \end{minipage}
  \hspace{.2cm}
  \begin{minipage}{0.5\textwidth}
  \centering
  \includegraphics[width=\hsize]{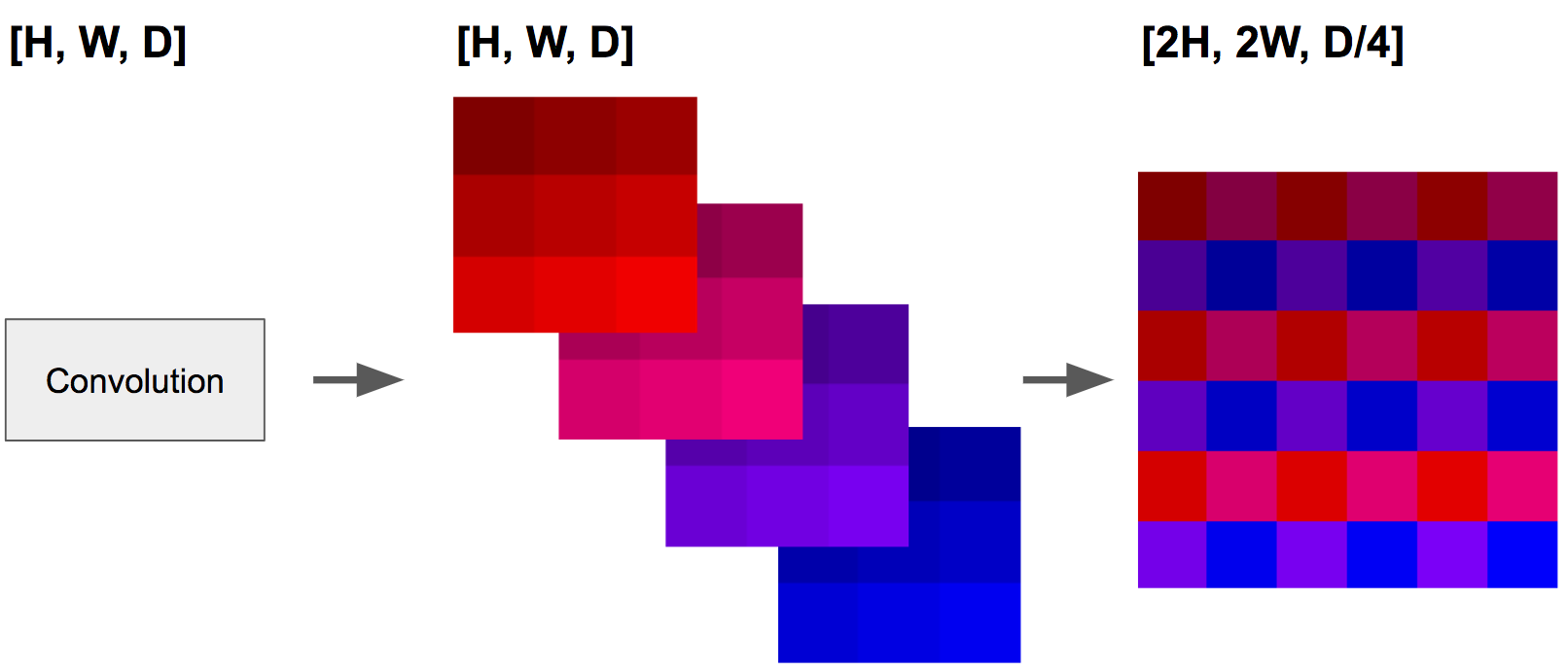}
  \caption{Visual illustration of Depth-to-Space. Fist a normal convolution is applied, keeping both the resolution and the number of channels. Then features of each four consecutive channels are re-arranged spatially into a single channel, effectively reducing the number of channels by a factor 4.}
  \label{fig:depthtospace}
  \end{minipage}
  \begin{minipage}{\textwidth}
  \vspace{.8cm}
    \centering
  \includegraphics[width=0.85\hsize]{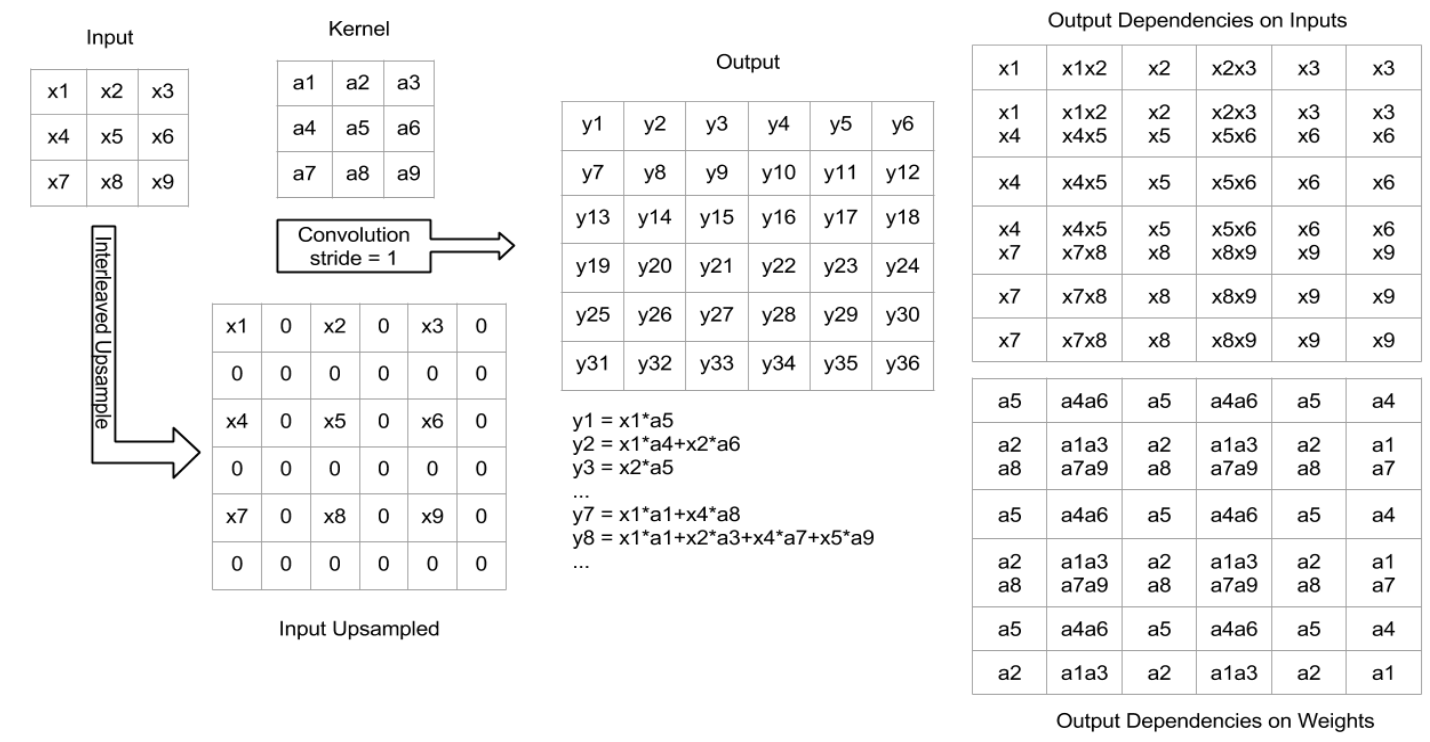}
  \caption{Spatial dependency for the outputs of a Transposed Convolution with kernel size 3 and stride 2.}
  \label{fig:transpose}
  \end{minipage}
  \begin{minipage}{\textwidth}
  \vspace{.8cm}
    \centering
  \includegraphics[width=0.85\hsize]{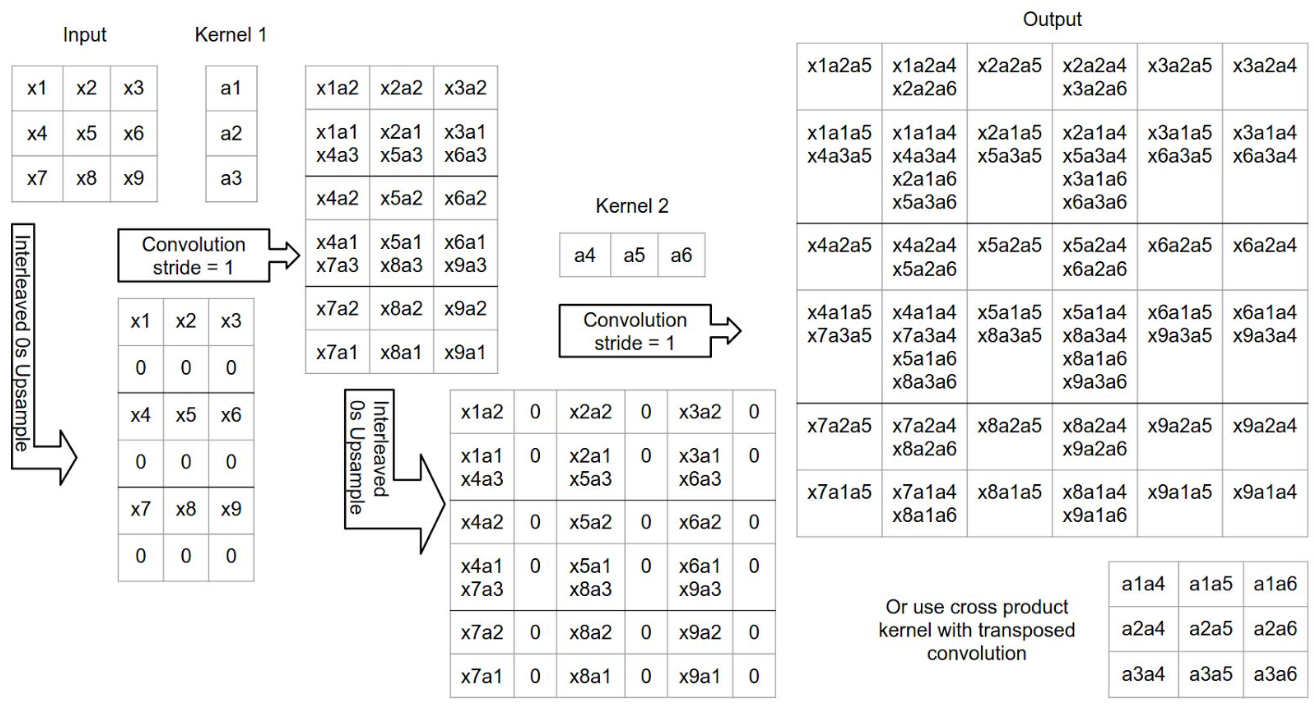}
  \caption{Spatial dependency for the outputs of a Decomposed Transposed Convolution with kernel size 3 and stride 2.}
  \label{fig:dectranspose}
\end{minipage}
\end{figure*}

\mytoppar{Transposed Convolution.}
Transposed convolutions are the most commonly used upsampling layers and are also sometimes referred to as `deconvolution' or `upconvolution' \citep{long2015fully, dumoulin2016guide, zeiler2010deconvolutional,dosovitskiy2016inverting, fischer2015flownet, DBLP:journals/corr/RonnebergerFB15, laina2016deeper}.
A transposed convolution can be seen as a reversed convolution in the sense of how the input and output are related to each other. However, it is not an inverse operation, since calculating the exact inverse is an under-constrained problem and therefore ill-posed. Transposed convolution is equivalent to interleaving the input features with 0's and applying a standard convolutional operation. The transposed convolution is illustrated in Fig.~\ref{fig:transpose_visual} and~\ref{fig:transpose}. \\

\mypar{Decomposed Transposed Convolution.}
Whereas decom\-posed transposed convolution is similar to the transposed convolution, conceptually it splits the main convolution operation into multiple low-rank convolutions. For images, it simulates a 2D transposed convolution using two 1D convolutions (Fig.~\ref{fig:dectranspose}). Regarding possible feature transformations, decomposed transposed convolution is strictly a subset of regular transposed convolution. As an advantage, the number of trainable parameters is reduced (Tab.~\ref{tab:comparison}).

Decomposed transposed convolution was successfully applied in the inception architecture \citep{DBLP:journals/corr/SzegedyVISW15}, which achieved state of the art results on ILSVRC 2012 classification~\citep{ILSVRC15}. It was also used to reduce the number of parameters of the network in \citep{DBLP:journals/corr/AlvarezP16}. \\

\mypar{Conv + Depth-To-Space.}
Depth-to-space \citep{shi2016real} (also called subpixel convolution) shifts the feature channels into the spatial domain as illustrated in Fig.~\ref{fig:depthtospace}. Depth-to-space preserves perfectly all floats inside the high dimensional representation of the image, as it only changes their placement. The drawback of this approach is that it introduces alignment artifacts. To be comparable with other upsampling layers which have learnable parameters, before performing the depth-to-space transformation we apply a convolution with four times more output channels than for other upsampling layers.

\mypar{Bilinear Upsampling + Convolution.} Bilinear Interpolation is another conventional approach for upsampling
the spatial resolution. To be comparable to other methods, we assume there is additional
convolutional operation applied after the upsampling. The drawback of this strategy is that it is
both memory and computationally intensive: bilinear interpolation increases the feature size
quadratically while keeping the same amount of ``information'' as measured in the number of floats. Because bilinear upsampling is followed by a convolution, the resulting upsampling method is four times more expensive than a transposed convolution. \\

\mypar{Bilinear Upsampling + Separable Convolution.}
Separable convolution was used to build a simple and homogeneous network architecture
\citep{DBLP:journals/corr/Chollet16a} which achieved superior results to inception-v3
\citep{DBLP:journals/corr/SzegedyVISW15}. A separable convolution consists of two operations, a per
channel convolution and a pointwise convolution with $1 \times 1$ kernel which mixes the channels. 
To increase the spatial resolution, before separable convolution, we apply bilinear upsampling in our experiments. \\

\subsubsection{Bilinear Additive Upsampling}
\begin{figure*}
  \begin{minipage}{0.5\textwidth}
  \centering
  \includegraphics[width=\hsize]{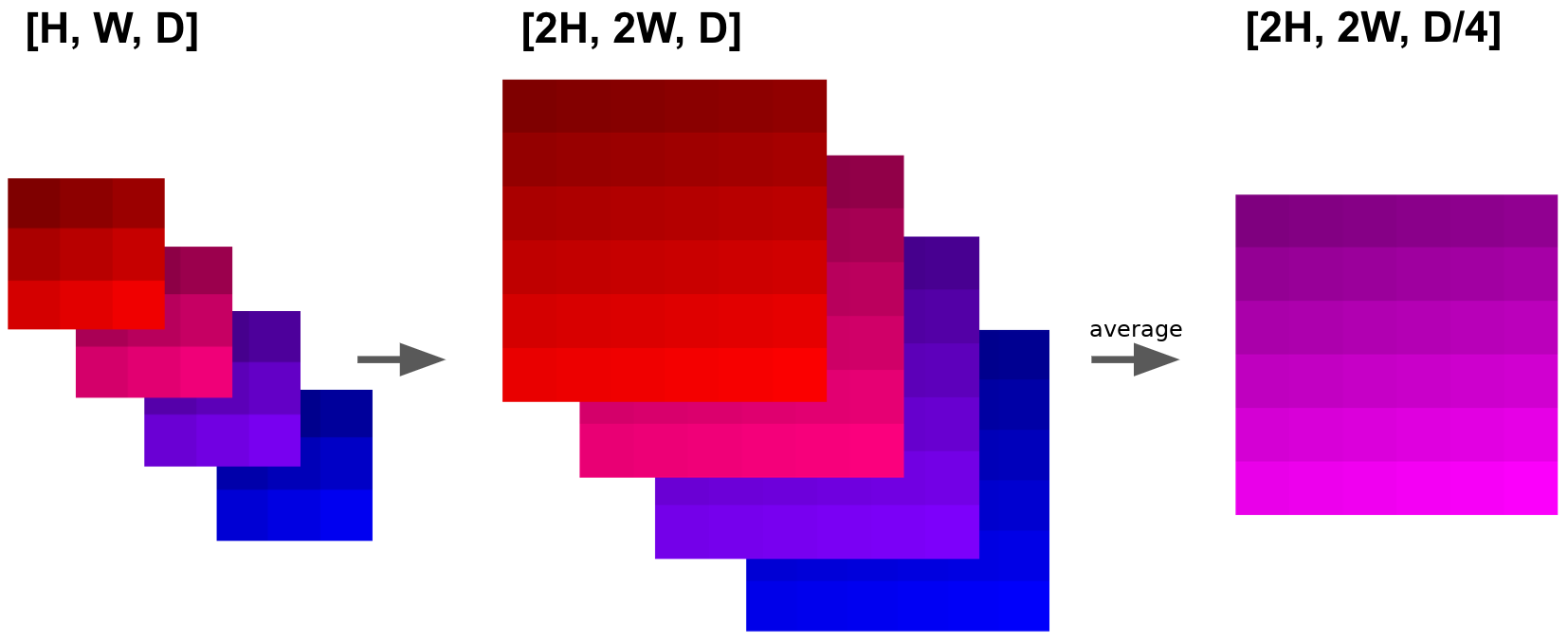}
  \caption{Visual illustration of our bilinear additive upsampling. We upsample all features and take the average over each four consecutive channels. For our upsampling layer this operation is followed by a convolution (not visualized).}
  \label{fig:resupsampling}
  \end{minipage}
  \hspace{.2cm}
  \begin{minipage}{0.5\textwidth}
  \centering
  \includegraphics[width=\hsize]{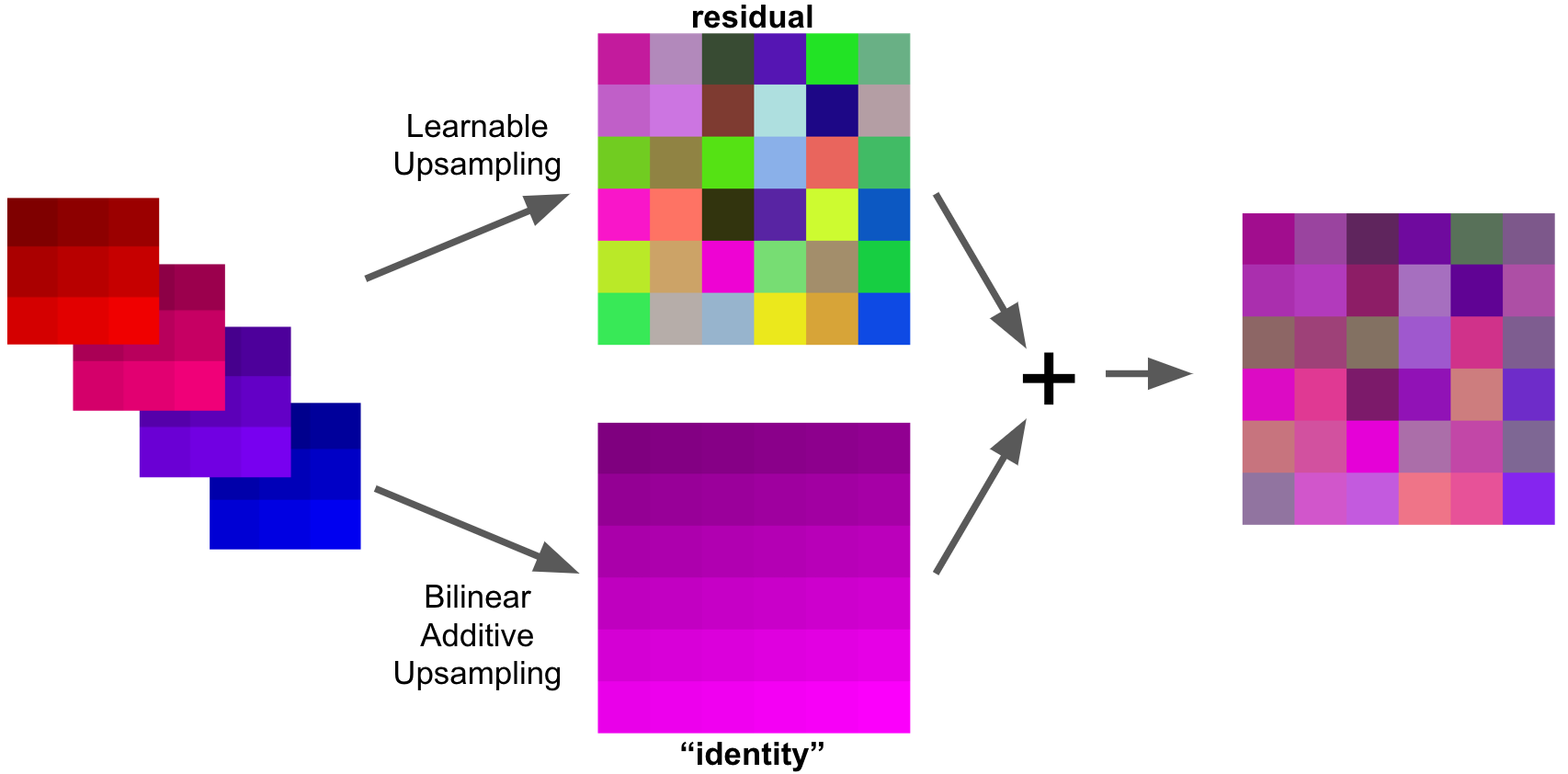}
  \caption{Visual illustration of our residual connection for upsampling layers. The ``identity'' layer is the sum of each four consecutive upsampled layers in order to get the desired number of channels and resolution.}
  \label{fig:residualconnection}
  \end{minipage}
\end{figure*}

To overcome the memory and computational problems of bilinear upsampling, we introduce a new upsampling layer: bilinear additive upsampling. In this layer, we propose to do bilinear upsampling as before, but we also add every $N$ consecutive channels together, effectively reducing the output by a factor $N$. This process is illustrated in Fig.~\ref{fig:resupsampling}. Please note that this process is deterministic and has zero tunable parameters (similarly to depth-to-space upsampling, but does not cause alignment artifacts). Therefore, to be comparable with other upsampling methods we apply a convolution after this upsampling method. In this paper, we choose $N$ in such a way that the final number of floats before and after bilinear additive upsampling is equivalent (we upsample by a factor 2 and choose $N=4$), which makes the cost of this upsampling method similar to a transposed convolution.

\subsection{Skip Connections and Residual Connections}

\subsubsection{Skip Connections}
Skip connections have been successfully used in many decoder architectures \citep{DBLP:journals/corr/LinMS016,DBLP:journals/corr/RonnebergerFB15,DBLP:journals/corr/PinheiroLCD16,DBLP:journals/corr/LinDGHHB16,DBLP:journals/corr/KendallBC15}. This method uses features from the encoder in the decoder part of the same spatial resolution, as illustrated in Fig.~\ref{fig:encdec}. For our implementation of skip connections, we apply the convolution on the last layer of encoded features for a given spatial resolution and concatenate them with the first layer of decoded features (Fig.~\ref{fig:encdec}).

\subsubsection{Residual Connections for decoders}
Residual connections \citep{DBLP:journals/corr/HeZRS15} have been shown to be beneficial for a variety of tasks. However, residual connections cannot be directly applied to upsampling methods since the output layer has a higher spatial resolution than the input layer and a lower number of feature channels. In this paper, we introduce a transformation which solves both problems.

In particular, the bilinear additive upsampling method which we introduced above (Fig.~\ref{fig:resupsampling}) transforms the input layer into the desired spatial resolution and number of channels without using any parameters. The resulting features contain much of the information of the original features. Therefore, we can apply this transformation (this time without doing any convolution) and add its result to the output of any upsampling layer, resulting in a residual-like connection. We present the graphical representation of the upsampling layer in Fig.~\ref{fig:residualconnection}. We demonstrate the effectiveness of our residual connection in Section~\ref{cpt:results}.

\begin{table*}[t]
\vspace{0.2cm}
\footnotesize
\begin{center}
\begin{tabular}{| c | c | c | c |}  \hline
Upsampling method                          & \# of parameters & \# of operations          & Comments \\ \hline
Transposed                                       & $whIO$               & $whWHIO$                & \\ 
Decomposed Transposed                 & $(w+h)IO$          & $(w+h)WHIO$            & Subset of the Transposed method \\  
Conv + Depth-To-Space                    & $whI(4O)$          & $whWHI(4O)$            & \\  
Bilinear Upsampling + Conv              & $whIO$              & $wh(2W)(2H)IO$        & \\
Bilinear Upsampling + Separable      & $whI + IO$         & $(2W)(2H)I(wh + O)$ & \\  
Bilinear Additive Upsampling + Conv & $whIO$             & $wh(2W)(2H)(I/4)O$   & \\ 
\hline
\end{tabular}
\end{center}
\caption{Comparison of different upsampling methods. $W,H$ - feature width and height, $w,h$ - kernel width and height, $I,O$ - number of channels for input and output features.}
\label{tab:comparison}
\end{table*}

\section{Tasks and Experimental Setup}
\subsection{Classification}
\subsubsection{Semantic segmentation}

We evaluate our approach on the standard PASCAL VOC-2012 dataset \citep{pascal-voc-2012}. We use both
the training dataset and augmented dataset \citep{hariharan2011semantic} which together consist of
10,582 images. We evaluate on the VOC Pascal 2012 validation dataset of 1,449 images.  We follow
a similar setup to Deeplab-v2 \citep{DBLP:journals/corr/ChenPK0Y16}. As the encoder network, we 
use ResNet-101 with stride 16. We replace the first 7x7 convolutional layer with three
3x3 convolutional layers. We use [1, 2, 4] atrous rates in the last three convolutional layers of the
block5 in Resnet-101 as in \citep{DBLP:journals/corr/ChenPSA17}. We use batch normalization, 
have single image pyramid by using [6, 12, 18] atrous rates together with global feature pooling. We initialize our model with the pre-trained weights on
ImageNet dataset with an input size of $513 \times 513$. Our decoder upsamples four times by a
factor 2 with a single 3x3 convolutional layer in-between.  We train the network with stochastic
gradient descent on a single GPU with a momentum of 0.9 and batch size 16. We start from learning rate
0.007 and use a polynomial decay with the power 0.9 for 30,000 iterations. We apply L2 regularization
with weight decay of $0.0001$. We augment the dataset by rescaling the images by a
random factor between 0.5 and 2.0. During evaluation, we combine the prediction from
multiple scales [0.5, 0.75, 1.0, 1.25, 1.5, 1.75] and the flipped version of the input image.  We
train the model using maximum likelihood estimation per pixel (softmax cross entropy) and use mIOU
(mean Intersection Over Union) to benchmark the models.

\subsubsection{Instance boundaries detection}
For instance-wise boundaries, we use PASCAL VOC 2012 segmentation \citep{pascal-voc-2012}. This
data\-set contains 1,464 training and 1,449 validation images, annotated with contours for 20 object classes for all instances. The dataset was originally designed for semantic segmentation. Therefore, only interior object pixels are marked, and the boundary location is recovered from the segmentation mask. Similar to \citep{uijlings2015situational} and \citep{DBLP:journals/corr/KhorevaBO0S15}, we consider only object boundaries without distinguishing semantics, treating all 20 classes as one.

As an encoder or feature extractor we use ResNet-50 with stride 8 and atrous convolution. We initialize from pre-trained ImageNet weights. The input to the network is of size $321 \times 321$. The spatial resolution is reduced to $41 \times 41$, after which we use $3$ upsampling layers, with an additional convolutional layer in-between, to make predictions in the original resolution.

During training, we augment the dataset by rescaling the images by a random factor between 0.5 and 2.0 and random cropping. We train the network with asynchronous stochastic gradient descent for 40,000 iterations with a momentum of 0.9. We use a learning rate of 0.0003 with a polynomial decay of power 0.99. We apply L2 regularization with a weight decay of 0.0002. We use a batch size of $5$. We use sigmoid cross entropy loss per pixel (averaged across all pixels), where $1$ represents an edge pixel, and 0 represents a non-edge pixel.

We use two measures to evaluate edge detection: f-measure for the best-fixed contour threshold across the entire dataset and average precision (AP). During the evaluation, predicted contour pixels within three pixels from ground truth pixels are assumed to be correct \citep{MartinFTM01}.

\subsection{Regression}

\subsubsection{Human keypoints estimation}\label{sec:human_keypoints}
We perform experiments on the MPII Human Pose data\-set~\citep{andriluka14cvpr} which consists of around 25k images with over 40k annotated human poses in terms of keypoints of the locations of seven human body parts. The images cover a wide variety of human activities. Since test annotations are not provided and since there is no official train and validation split, we make such split ourselves (all experiments use the same split). In particular, we divide the training set randomly into 80\% training images and 20\% validation images.
Following~\citep{andriluka14cvpr}, we measure the Percentage of Correct Points with a matching threshold of 50\% of the head segment length (PCKh), and report the average over all body part on our validation set.

For the network, we re-implement the stacked hourglass network~\citep{DBLP:journals/corr/NewellYD16} with a few modifications. In particular, we use as input a cropped the area around the center of human of size 353x272 pixels. We resize predictions from every hourglass subnetwork to their input resolution using bilinear interpolation and then we apply a Mean Squared Error (MSE) loss. The target values around each keypoint are based on a Gaussian distribution with a standard deviation of 6 pixels in the original input image, which we rescale such that its highest value is 10 (which we found to work better with MSE). We do not perform any test-time data augmentation such as horizontal flipping.

\subsubsection{Depth prediction}
We apply our method to the NYUDepth v2 dataset by~\citep{Silberman:ECCV12}. We train on the entire NYU\-Depth v2 raw data distribution, using the official split. There are $209,822$ train and $187,825$ test images. We ignore the pixels that have a depth below a threshold of 0.3 meters in both training and test, as these reads are not reliable in the Kinect sensor.

As for the encoder network, we use ResNet-50 with stride $32$, initialized with pre-trained ImageNet weights. We use an input size of $304 \times 228$ pixels. We apply $1x1$ convolution with 1024 filters. Following \citep{laina2016deeper}, we upsample four times with a convolutional layer in between to get back to half of the original resolution. Afterwards, we upsample with bilinear interpolation to the original resolution (without any convolutions).

We train the network with asynchronous stochastic gradient descent on $3$ machines with a momentum of 0.9, batch size 16, starting from a learning rate of $0.001$, decaying by $0.92$ every $72926$ steps and train for $640,000$ iterations. We apply L2 regularization with weight decay $0.0005$. We augment the dataset through random changes in brightness, hue, and saturation, through random color removal and through mirroring.

For depth prediction, we use the reverse Huber loss following \citep{laina2016deeper}. 
\begin{equation}
Loss(y, \hat{y}) = \begin{cases}
                    |y-\hat{y}|   \quad for \quad |y-\hat{y}| <= c \\
                    |y-\hat{y}|^2 \quad for \quad |y-\hat{y}| > c
                   \end{cases}
\end{equation}
\begin{equation}
c = \frac{1}{5} \max \limits_{(b,h,w) \in [1\ldots Batch][1\ldots Height][1 \ldots Width]} | y_{b,h,w} - \hat{y_{b,h,w}} |
\end{equation}
The reverse Huber loss is equal to the L1 norm for $x \in [-c, c]$ and equal to L2 norm outside this range. In every gradient descent step, c is set to $20\%$ of the maximal pixel error in the batch.

For evaluation, we use the metrics from \citep{DBLP:journals/corr/EigenF14}, i.e., mean relative error, root mean squared error, root mean squared log error, the percentage of correct prediction within three relative thresholds: $1.25$, $1.25^2$ and $1.25^3$.

\subsubsection{Colorization}
We train and test our models on the ImageNet dataset \citep{ILSVRC15}, which consists of $1,000,000$ training images and $50,000$ validation images.

For the network architecture, we follow \citep{iizuka2016let}, where we swap their original bilinear
upsampling method with the methods described in Section~\ref{sec:method}. In particular, these are three upsampling steps of factor 2.

This model combines joint training of image classification and colorization, where we are mainly interested in the colorization part. We resize the input image to $224 \times 224$ pixels. We train the network for $30,000$ iterations using the Adam optimizer with a batch size of 32 and we fix the learning rate to 1.0. We apply L2 regularization with a weight decay of 0.0001. During training, we randomly crop and randomly flip the input image. For the skip connections, we concatenate the decoded features with the feature extractor applied to the original input image (as there are two encoder networks employed on two different resolutions).

As loss function we use the averaged L1 loss for pixel-wise color differences for the colorization part, and a softmax cross entropy loss for the classification part.
\begin{equation}
Loss(y, \hat{y}, y_{cl}, \hat{y_{cl}}) = 10 | y - \hat{y} | - y_{cl} \log \hat{y_{cl}}
\end{equation}
Color predictions are made in the YPbPr color space (luminance, blue - luminance, red - luminance). The luminance is ignored in the loss function during both training and evaluation as is it provided by the input greyscale image. The output pixel value targets are scaled to the range [0,1]. $y_cl$ is the one hot encoding of the predicted class label and $\hat{y_{cl}}$ are the predicted classification logits.

To evaluate colorization we follow \citep{zhang2016colorful}. We compute the average root mean squared error between the color channels in the predicted and ground truth pixels. Then, for different thresholds for root mean squared errors, we calculate the accuracy of correctly predicted colored pixels within given range. From these we compute the Area Under the Curve (AUC) \citep{zhang2016colorful}. Additionally, we calculate the top-1 and top-5 classification accuracy for the colorized images on the ImageNet~\cite{ILSVRC15} dataset, motivated by the assumption that better recognition corresponds to more realistic images.

\subsubsection{Super resolution}\label{sec:super_resolution}
For super-resolution, we test our approach on the CelebA dataset, which consists of $167,483$ training images and $29,249$ validation images \citep{liu2015faceattributes}. We follow the setup from \citep{yu2016ultra}: the input images of the network are $16 \times 16$ images, which are created by resizing the original images. The goal is to reconstruct the original images which have a resolution of $128 \times 128$.

The network architecture used for super-resolution is similar to the one from \citep{DBLP:journals/corr/KimLL15b}. We use six ResNet-v1 blocks with 32 channels after which we upsample by a factor of 2. We repeat this three times to get to a target upsampling factor of $8$. On top of this, we add 2 pointwise convolutional layers with 682 channels with batch normalization in the last layer. Note that in this problem there are only operations which keep the current spatial resolution or which upsample the representation. We train the network on a single machine with 1 GPU, batch size 32, using the RMSProp optimizer with a momentum of 0.9, a decay of 0.95 and a batch size of 16. We fix the learning rate at 0.001 for 30000 iterations. We apply L2 regularization with weight decay 0.0005. The network is trained from scratch.

As loss we use the averaged L2 loss between the predicted residuals $\hat{y}$ and actual residuals $y$. The ground truth residual $y$ in the loss function is the difference between original $128 \times 128$ target image and the predicted upsampled image. All target values are scaled to [-1,1].
We evaluate performance using standard metrics for super-resolution: PSNR and SSIM.

\subsection{Synthesis}

\subsubsection{Generative Adversarial Networks}\label{sec:gans}
To test our decoders in the generator network for Generative Adversarial
Networks~\citep{goodfellow14nips} we follow the setup from \citep{DBLP:journals/corr/abs-1711-10337}.  We benchmark different decoders (generators) on the Cifar-10 dataset which consists of 50000 training and 10000 testing 32x32 color images. Additionally, we also test our approach on the CelebA dataset described in Sec.~\ref{sec:super_resolution}.

For Cifar-10 we use the GAN architecture with spectral normalization of~\cite{miyato2018spectral}. For CelebA, we use the InfoGan architecture~\citep{DBLP:conf/nips/ChenCDHSSA16}. For both networks, we replace the transposed convolutions in the generator with our upsampling layers and add 3x3 convolutional layers between them.  In the discriminator network, we do not use batch normalization. We train the model with batch size 64 through 200k iterations on a single GPU using the Adam optimizer. We perform extensive hyperparameter search for each of the studied decoder architectures: We vary the learning rate from a logarithmic scale between [0.00001, 0.01], we vary the beta1 parameter for the Adam optimizer from range [0, 1], and the $\lambda$ used in the gradient penalty term (Eq.~\eqref{eq:gan_discriminator}) from a logarithmic scale in range [-1, 2]. We try both 1 and 5 discriminator updates per generator update.

As the discriminator and generator loss functions we are using the Wasserstein GAN Loss with Gradient Penalty \citep{DBLP:conf/nips/GulrajaniAADC17}:
 
\begin{equation}\label{eq:gan_discriminator}
\begin{aligned}
\mathcal{L}_D = & - \mathbb{E}_{x \sim p_d} [D(x)] \quad + \quad \mathbb{E}_{\hat{x} \sim p_g} [D(\hat{x})] \quad + \\
                           & \lambda \mathbb{E}_{\hat{x} \sim p_g} [ (|| \nabla D ( \alpha x + (1 -
                           \alpha \hat{x})) ||_2 - 1)^2 ] \\ \\ 
\end{aligned}
\end{equation}
\begin{equation}
\begin{aligned}
\mathcal{L}_G = & - \mathbb{E}_{\hat{x} \sim p_g} [D(\hat{x})] 
\end{aligned}
\end{equation}

where $p_d$ is data distribution, $p_g$ is the generator output distribution, and $D$ is the output of the discriminator. $\alpha$ is uniformly sampled in every iteration from range [0, 1].

We evaluate the performance of the generator architectures using the Frechet Inception Distance (FID) \citep{DBLP:conf/nips/HeuselRUNH17}.

\section{Results}\label{cpt:results}

\begin{table*}
  \centering
  \subtable[Results without residual connections.]{\includegraphics[width=1.0\vsize,angle=90]{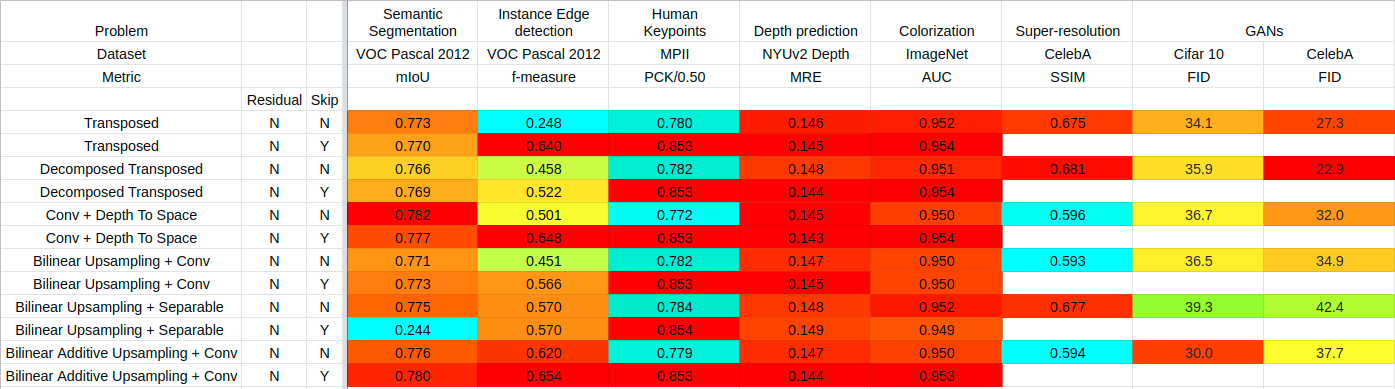}\label{tab:results_table_1}}
  \hspace{3cm}
  \subtable[Results with residual connections.]{\includegraphics[width=1.0\vsize,angle=90]{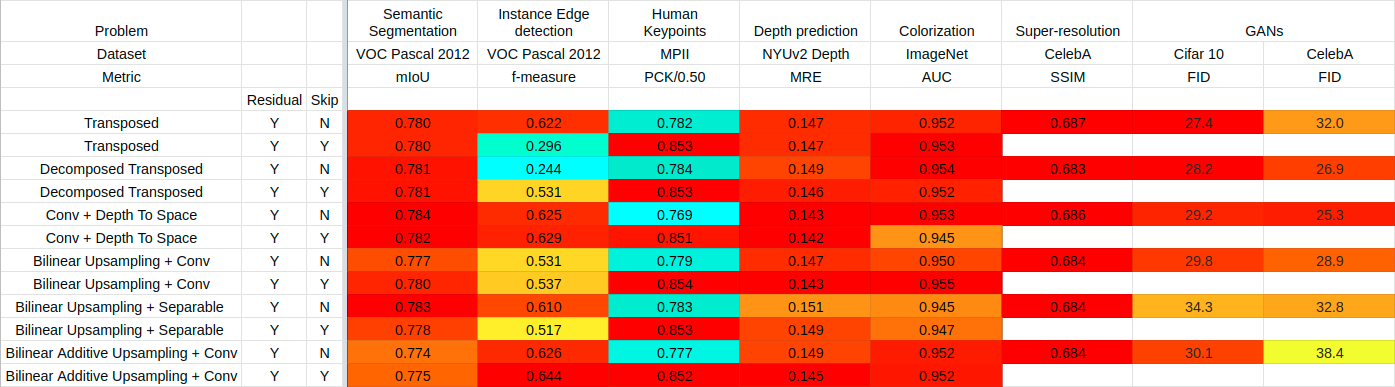}\label{tab:results_table_2}}
\caption{Our main results comparing a variety of decoders on seven machine vision problems. The colors represent relative performance: red means top performance, yellow means reasonable performance, blue and green means poor performance.}
\end{table*}

We first compare the upsampling types described in Section~\ref{sec:upsampling} without residual connections (Tab.~\ref{tab:results_table_1}). Next, we discuss the benefits of adding our residual connections (Tab.~\ref{tab:results_table_2}).
Since all evaluation metrics are highly correlated, this table only reports a single metric per problem. A table with all metrics can be found in the supplementary material.

\begin{table*}[t]
\footnotesize
\begin{center}
\begin{tabular}{| l | l | l | l | l | }  \hline
Method                          & Measure   & Our method & Recent work & Recent work \\ \hline \hline
Semantic segmentation & mIoU        & 0.77           &  0.76 & \cite{DBLP:journals/corr/ChenPK0Y16}  \\ \hline
Instance boundaries detection              & f-measure & 0.63             & 0.62    & \cite{DBLP:journals/corr/KhorevaBO0S15} \\ \hline
Human keypoints estimation & PCKh & $0.85^*$ & $0.88^*$ & \cite{DBLP:journals/corr/NewellYD16} \\ \hline
Depth prediction           & MRE          & 0.15           & 0.15  & \cite{laina2016deeper} \\ \hline
Colorization                   & AUC          & 0.95           & 0.90  & \cite{zhang2016colorful} \\ \hline
Super-resolution            & SSIM        & 0.68             & 0.70    & \cite{yu2016ultra} \\ \hline
GANs on Cifar 10            & FID           & 30           & 53    & \cite{DBLP:journals/corr/abs-1711-10337} \\ \hline
\end{tabular}
\end{center}
\caption{Comparison of our bilinear additive upsampling + conv + res results with other methods from the literature. Comparing semantic segmentation results with \citep{DBLP:journals/corr/ChenPK0Y16}, we did not use MS COCO detection dataset for pretraining and multiscale training. The performance of the depth prediction task is compared on all the scene frames from the test dataset and not on small subset as in \citep{laina2016deeper}, therefore we report the numbers of our reimplementation of the method. $^*$Numbers on human keypoints estimation are not directly comparable since we use a slightly smaller training set and evaluate on a different random validation set. We see that for all tasks we achieve good results.}
\label{tab:other_results}
\end{table*}

\subsection{Results without residual-like connections.}
For semantic segmentation, the depth-to-space transformation is the best upsampling method.
For instance edge detection and human keypoints estimation, the skip-layers are necessary to get good results. For instance edge detection, the best performance is obtained by transposed, depth-to-space, and bilinear additive upsampling. For human keypoints estimation, the hourglass network uses skip-layers by default and all types of upsampling layers are about as good.
For both depth prediction and colorization, all upsampling methods perform similarly, and the specific choice of upsampling matters little.
For super-resolution, networks with skip-layers are not possible because there are no encoder modules which high-resolution (and relatively low-seman\-tic) features. Therefore, this problem has no skip-layer entries.
Regarding performance, only all transposed variants perform well on this task; other layers do not.
Similarly to super-resolution, GANs have no encoder and therefore it is not possible to have skip connections. Separable convolutions perform very poorly on this task. There is no single decoder which performs well on both Cifar 10 and CelebA datasets: bilinear additive upsampling performs well on Cifar 10, but poorly in CelebA. In contrast, decomposed transposed performs well on CelebA, but poorly on Cifar 10.

Generalizing over problems, we see that no single upsampling scheme provides
consistently good results across all the tasks.

\subsection{Results with residual-like connections}
Next, we add our residual-like connections to all upsampling methods. Results are presented in 
Tab.~\ref{tab:results_table_2}. For the majority of combinations, we see that adding
residual connections is beneficial. In fact, for semantic segmentation, depth prediction,
colorization, and super-resolution, adding residual connections results in consistently high performance across decoder types. For instance edge detection, transposed convolutions, depth-to-space, and bilinear additive upsampling work well when no skip connections are used. The only task which is unaffected by adding residual connections is human keypoints estimation. This is because there are already residual connections over each hourglass in the stacked hourglass network; each hourglass can be seen as refining the previous predictions by estimating residuals. Adding even more residual connections to the network does not help.

To conclude: (1) Residual-like connections almost always improve performance. (2) With residual
connections, we can identify three consistently good decoder types: transposed convolutions,
depth-to-space, and bilinear additive upsampling.

\begin{figure*}
  \centering
  \begin{tabular}{lccc}
    & Depth-to-Space & Transposed & Bilinear Additive Upsampling \\
    no residual & \includegraphics{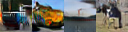} & \includegraphics{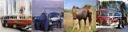} & \includegraphics{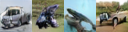} \\
    with residual & \includegraphics{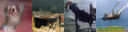} & \includegraphics{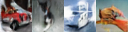} & \includegraphics{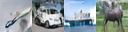} \\
  \end{tabular}
  \caption{Visualizations of GANs}
  \label{fig:gans}
\end{figure*}

\subsection{Decoder Artifacts}

The work of~\citep{odena2016deconvolution} identified checkerboard artifacts in GANs by giving qualitative
examples followed by an analysis of the decoder as to why these artifacts appear. In particular,
they showed that transposed convolutions have ``uneven overlap'', which means that convolutions at different places in the image operate on a different number of features, as is also demonstrated in Fig.~\ref{fig:transpose} (output dependencies on weights). Of course, this observation extends to all transposed convolution variants. We also note that a similar analysis holds for the depth-to-space decoder, where our explanation in Fig.\ref{fig:depthtospace} visually shows how the artifacts arise.

\begin{figure}
\begin{center}
\includegraphics[width=\hsize]{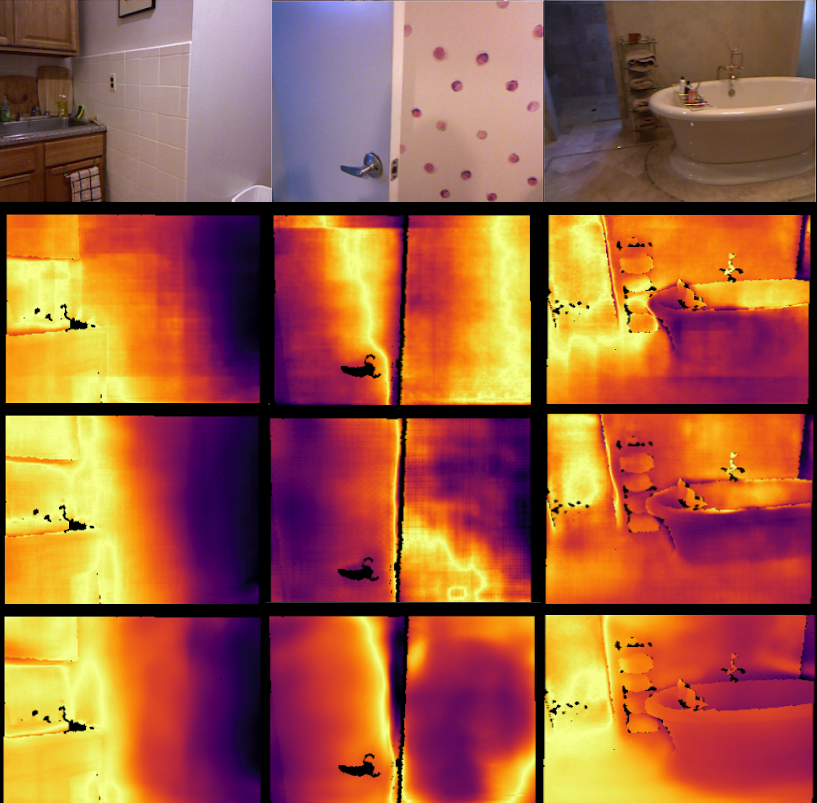}
\caption{Input image and errors for three different models. Top row: input image. Second row: Transposed. Third row: Depth-to-Space. Fourth row: Bilinear additive upsampling. We observe that our proposed upsampling method produces smoother results with less arifacts.}
\label{fig:qual_artifacts}
\end{center}
%\vspace{-0.8cm}
\end{figure}

We can indeed observe artifacts by looking at a few qualitative results:
Fig.~\ref{fig:gans} and Fig.~\ref{fig:qual_artifacts} show several output results of respectively GANs and the depth prediction task. We show outputs for transposed
convolutions, depth-to-space, and our new bilinear additive upsampling method. For GANs, when not having residual connections, depth-to-space upsampling frequently results in artifacts, as shown in the two left-most images. For transposed convolutions there are fewer and generally more subtle artifacts (see the right-most image), while we could not find these for our bilinear additive upsampling method. When adding residual connections, the depth-to-space and transposed methods stop producing visible artifacts. Similarly, for the depth-prediction task, without residual connections, depth-to-space and transposed convolutions result in clearly visible artifacts, while our method yields much better looking results.

For the depth-prediction problem, we can also directly measure a certain type of artifact. To do this, we
observe that object surfaces are mostly smooth. This means that if we move over the pixels of an
object in x- or y-direction, we expect the depth on these objects to either monotonically increase
or decrease. In particular, we consider triplets of pixels in either horizontal or vertical
direction on surfaces which face the camera (ground-truth depth difference between adjacent pixels
is smaller than 1 cm) and for which the ground truth is monotonically increasing/decreasing. We then
measure the percentage of triplets in which the predictions are non-monotonic. Results are presented
in Tab.~\ref{tab:artefacts}, where a higher percentage means more artifacts.

\begin{table}[th]
\footnotesize
\begin{center}
\begin{tabular}{| l | l | l | l | l | }  \hline
Upsampling layer & Skip & Res & artifacts & artifacts \\ 
                             &        &         & x-axis    & y-axis     \\ \hline \hline
Transposed          & N     & N     & 27.6\% & 24.8\% \\ 
Transposed          & Y     & N     & 30.6\% & 23.5\% \\ 
Transposed          & N     & Y     & 23.1\% & 21.9\% \\
Transposed          & Y     & Y     & 22.8\% & 22.8\% \\ \hline
Dec. Transposed  & N    & N     & 27.0\% & 24.1\% \\
Dec. Transposed  & Y    & N     & 33.4\% & 24.0\% \\
Dec. Transposed  & N    & Y     & 25.5\% & 23.9\% \\
Dec. Transposed  & Y     & Y     & 24.5\% & 24.1\% \\ \hline
Depth-To-Space    & N    & N     & 28.4\% & 26.9\% \\
Depth-To-Space    & Y    & N     & 26.3\% & 23.2\% \\
Depth-To-Space    & N    & Y     & 25.6\% & 23.7\% \\
Depth-To-Space    & Y    & Y      & 24.2\% & 22.9\% \\ \hline
Bilinear Ups. + Conv. & N    & N      & 14.3\% & 14.6\% \\
Bilinear Ups. + Conv. & Y    & N       & 19.5\% & 19.5\% \\
Bilinear Ups. + Conv. & N    & Y       & 14.7\% & 14.8\% \\
Bilinear Ups. + Conv. & Y    & Y       & 17.4\% & 18.3\% \\ \hline
Bilinear Ups. + Sep.   & N    & N     & 13.8\% & 14.6\% \\
Bilinear Ups. + Sep.   & Y    & N      & 20.5\% & 20.3\% \\
Bilinear Ups. + Sep.   & N    & Y      & 17.3\% & 16.0\% \\
Bilinear Ups. + Sep.   & Y    & Y      & 19.7\% & 18.3\% \\ \hline
Bilinear Add. Ups. & N    & N      & 14.8\% & 15.2\% \\
Bilinear Add. Ups. & Y    & N       & 19.0\% & 17.7\% \\
Bilinear Add. Ups. & N    & Y       & 14.7\% & 15.0\% \\
Bilinear Add. Ups. & Y    & Y       & 17.5\% & 18.4\% \\ \hline
\end{tabular}
\end{center}
\caption{Analysis of the artifacts for depth regression for 1000 test images, totalling approximately 27 million triplets per experiment.}
\label{tab:artefacts}
\end{table}

Results confirm our qualitative observations and what we expect from the decoder
methods: the bilinear upsampling variants have significantly fewer artifacts than the transposed
variants and than depth-to-space. We also observe that for transposed variants and depth-to-space, residual connections help to reduce artifacts. We conclude that the bilinear upsampling methods are preferred for minimizing prediction artifacts.

\subsection{Nearest Neighbour and Bicubic Upsampling}
The upsampling layers in our study used bilinear upsampling if applicable. But one can also use nearest neighbour upsampling (e.g.~\cite{berthelot17arxiv,jia17arxiv,odena2016deconvolution}) or bicubic upsampling (e.g.~\cite{dong16tpami,hui16eccv,DBLP:journals/corr/KimLL15b,odena2016deconvolution}). We performed two small experiments to examine these alternatives.

In particular, on the Human keypoints estimation task (Sec.~\ref{sec:human_keypoints}) we started from the bilinear upsampling + conv method with residual connections and skip layers. Then we changed the bilinear upsampling to nearest neighbour and bicubic. Results are in Tab.~\ref{tab:keypoints_nn_cubic}. As can be seen, the nearest neighbour sampling is slightly worse than bilinear and bicubic, which perform about the same.

\begin{table}
  \footnotesize
  \centering
\begin{tabular}[htpb]{|l|c|c|c|}
  \hline
  & Res & Skip & PCK/0.50 \\
  \hline
NN Ups. + Conv & Y & Y & 0.846 \\
Bilinear Ups. + Conv & Y & Y & 0.854 \\
Bicubic Ups. Conv & Y & Y & 0.855 \\
  \hline
\end{tabular}
\caption{Comparison of Nearest Neighbour Upsamping, Bilinear Upsampling, and Bicubic Usampling on Human Keypoint Estimation.}
\label{tab:keypoints_nn_cubic}
\vspace{-0.2cm}
\end{table}

\begin{table}
  \footnotesize
  \centering
\begin{tabular}[htpb]{|l|c|c|c|}
  \hline
  & Res & Skip & FID \\
  \hline
NN Ups. + Conv & Y & N & 28.0 \\
Bilinear Ups. + Conv & Y & N & 29.8 \\
Bicubic Ups. Conv & Y & N & 28.1 \\
  \hline
\end{tabular}
\caption{Comparison of Nearest Neighbour Upsamping, Bilinear Upsampling, and Bicubic Usampling on generating synthetic images using GANs for Cifar 10.}
\label{tab:gan_nn_cubic}
\vspace{-0.2cm}
\end{table}

We also compared the same three decoder types to synthesize data using GANs (Sec.~\ref{sec:gans}) on Cifar 10 (without skip connections since these are not applicable). Results are in Tab.~\ref{tab:gan_nn_cubic}. As can be seen, nearest neighbour upsampling and bicubic have similar performance. This is in contrast to~\cite{odena2016deconvolution} who found that nearest neighbour upsampling worked better than bicubic upsampling. In~\cite{odena2016deconvolution} they suggested that their results could have been caused by the hyperparameters being optimized for nearest neighbour upsampling. Our result seems to confirm this. Bilinear upsampling performs slightly worse than the other two methods.

To conclude, in both experiments the differences between upsampling methods are rather small. Bicubic upsampling slightly outperforms the two others. Intuitively, this is the more accurate non-parametric upsampling method but one that has a higher computational cost.

\section{Conclusions}

This paper provided an extensive evaluation for different decoder types on a broad range of machine
vision applications. Our results demonstrate:
(1) Decoders matter: there are significant performance differences among different decoders
depending on the problem at hand.
(2) We introduced residual-like connections which, in the majority of cases, yield good improvements
when added to any upsampling layer.
(3) We introduced the bilinear additive upsampling layer, which strikes a right balance between the
number of parameters and accuracy. Unlike the other bilinear variants, it gives consistently good
performance across all applications.
(4) Transposed convolution variants and depth-to-space decoders have considerable prediction
artifacts, while bilinear upsampling variants suffer from this much less.
(5) Finally, when using residual connections, transposed convolutions, depth-to-space, and our
bilinear additive upsampling give consistently strong quantitative results across all problems.
However, since our bilinear additive upsampling suffers much less from prediction artifacts, it should be the upsampling method of choice.

\bibliographystyle{plainnat}     % basic style, author-year citations
\bibliography{shortstrings,egbib}
\afterpage{
\newpage~\newpage~
}

\newpage
\section{Appendix}

\begin{table*}
  \centering
  \subtable[Results without residual connections.]{\includegraphics[width=1.0\vsize,angle=90]{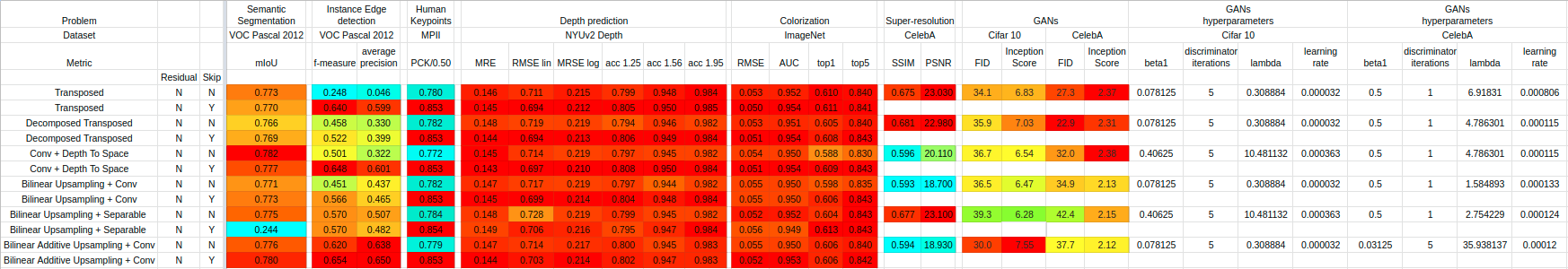}\label{tab:appendix_results_table_1}}
  \hspace{3cm}
  \subtable[Results with residual connections.]{\includegraphics[width=1.0\vsize,angle=90]{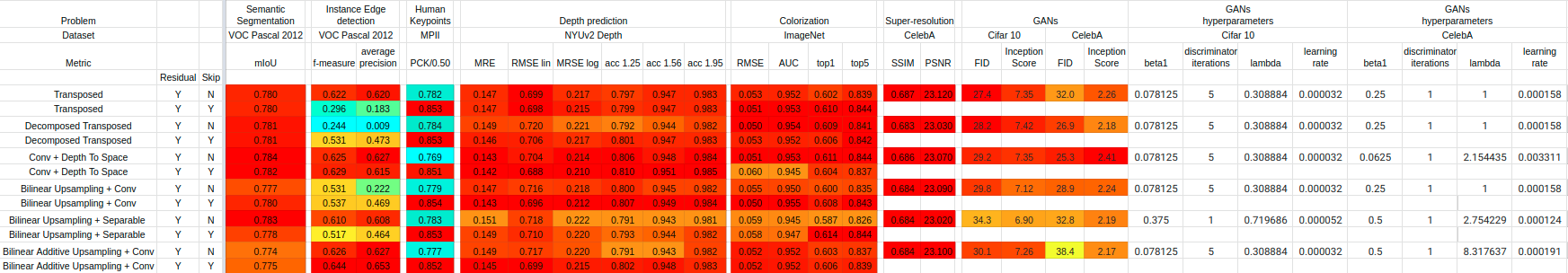}\label{tab:appendix_results_table_2}}
\caption{Our main results comparing a variety of decoders on seven machine vision problems. The colors represent relative performance: red means top performance, yellow means reasonable performance, blue and green means poor performance.}
\end{table*}

\end{document}